\documentclass{article}
\usepackage[preprint]{neurips_2024}

\usepackage[preprint]{neurips_2024}

\usepackage{comment} 
\usepackage{caption}
\usepackage{graphicx} 
\usepackage{subcaption} 
\usepackage{amsmath} 
\usepackage{multirow, multicol} 
\usepackage{amssymb} 
\usepackage{pifont}  
\usepackage{booktabs} 
\usepackage{algpseudocode}
\usepackage[linesnumbered,ruled,vlined]{algorithm2e}

\usepackage[utf8]{inputenc} 
\usepackage[T1]{fontenc}    
\usepackage{hyperref}       
\usepackage{url}            
\usepackage{booktabs}       
\usepackage{amsfonts}       
\usepackage{nicefrac}       
\usepackage{microtype}      
\usepackage{xcolor}         

\title{Multi-Bit Mechanism: A Novel Information Transmission Paradigm for Spiking Neural Networks}

\author{
    Yongjun Xiao, Xianlong Tian, Yongqi Ding, Pei He, Mengmeng Jing, Lin Zuo\textsuperscript{*} \\
    School of Information and Software Engineering\\
    University of Electronic Science and Technology of China\\
    \texttt{linzuo@uestc.edu.cn} \\
}

\begin{document}

\maketitle

\begin{abstract}

    Since proposed, spiking neural networks (SNNs) gain recognition for their high performance, low power consumption and enhanced biological interpretability. However, while bringing these advantages, the binary nature of spikes also leads to considerable information loss in SNNs, ultimately causing performance degradation. We claim that the limited expressiveness of current binary spikes, resulting in substantial information loss, is the fundamental issue behind these challenges. To alleviate this, our research introduces a multi-bit information transmission mechanism for SNNs. This mechanism expands the output of spiking neurons from the original single bit to multiple bits, enhancing the expressiveness of the spikes and reducing information loss during the forward process, while still maintaining the low energy consumption advantage of SNNs. For SNNs, this represents a new paradigm of information transmission. Moreover, to further utilize the limited spikes, we extract effective signals from the previous layer to re-stimulate the neurons, thus encouraging full spikes emission across various bit levels. We conducted extensive experiments with our proposed method using both direct training method and ANN-SNN conversion method, and the results show consistent performance improvements.
\end{abstract}

\section{Introduction}
In recent years, deep learning represented by artificial neural networks (ANNs) has been widely applied in many downstream tasks such as computer vision~\cite{{krizhevsky2012imagenet}} and game playing~\cite{silver2016mastering}, etc. However, these achievements are mostly built upon high computational budgets, which limits the application of ANNs on edge computing scenarios. 

As an alternative, spiking neural networks (SNNs) are proposed. SNNs transmit signals using discrete spikes with temporal characteristics, which is close to the information transmission mechanism of biological neurons, thus being regarded as the third generation of neural networks~\cite{maass1997networks}. Moreover, the sparse nature of the spikes makes the feature matrix sparse, reducing computational load, while their binary characteristics which transforms the multiply-and-accumulate (MAC) operations calculations into accumulate (AC) operations~\cite{roy2019towards}, simplifying the computation. All these, endows SNNs with high performance and low power consumption characteristics~\cite{diehl2015unsupervised}.

However, current SNNs also face challenges. Firstly, the forward process involves the conversion from full-precision membrane potentials to binary spikes, leading to significant information loss. Secondly, the network structures of existing SNNs mostly consist of simple modules where information is transmitted layer by layer, greatly reducing the network's expressive power. 

We claim that the roots of these two problems lie in the characteristics of the neuron structure and network architecture of SNNs, respectively. On one hand, in terms of neuron structure, existing SNNs, like ANNs, typically assume that there is only one synaptic connection between adjacent neurons. However, in biological neural networks, adjacent neurons often have multiple synapses. This may not be crucial for ANNs since they can transmit full-precision real values through a single synapse. But for SNNs, the presence of multiple synapses implies the possibility of transmitting multi-bit signals, which evidently helps to improve the precision of information transmission. Unfortunately, current SNNs simply adopt the single synapse setting like ANNs and thus lose the potential to transmit higher precision data. On the other hand, in terms of network architecture, the information flow in current SNNs is simply transmitted layer by layer. Such mechanism lacks the cross-layer information exchange existing in the biological neural networks, leading to the degenerated expressive power of SNNs. According to~\cite{thomson2003interlaminar}, projections from layer 3 to layer 5 exist in the adult mammalian neocortex.
\begin{figure}[h]
    \centering
    \includegraphics[width=\columnwidth]{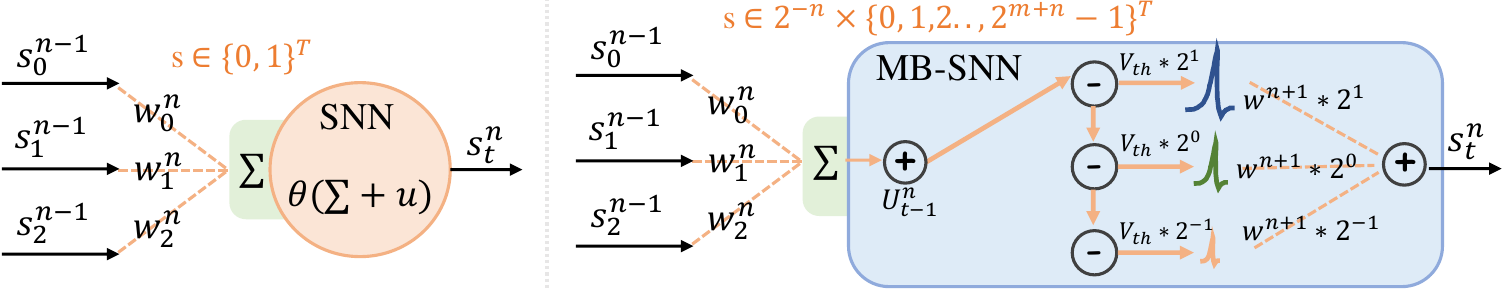}
    \caption{Neurons that emit single-bit spikes and neurons that emit multi-bit spikes.}
    \label{SNN_MBSNN}
\end{figure}

Based on the above considerations, we propose a multi-bit information transmission mechanism as shown in Fig.\ref{SNN_MBSNN} that expands the output of neurons from one bit to multiple bits without introducing extra parameters. This method significantly reduces information loss during the forward process of the network. Additionally, we simulate interlaminar connections. On one hand, this structure enriches the network's topological connections, enhancing the network's expressive power. On the other hand, through Efficient Channel Attention (ECA)~\cite{wang2020eca}, this structure extracts effective information from the previous layer of the network to stimulate the current neuron again, further compensating for the lost information. Our contributions are as follows:
\begin{itemize}
    \item We propose a multi-bit information transmission mechanism. This mechanism expands the output of spiking neurons from single bit to multiple bits. While preserving the low-energy advantages of SNNs, it significantly reduces information loss during the forward process. For SNNs, this represents a new paradigm of information transmission.
    \item We propose interlaminar connections. This approach further reduces information loss and enhances the network's expressive capability. Additionally, the sufficient retention of information makes the neurons more active, allowing the multi-bit mechanism to better realize its power.
    \item We conducted extensive experiments. The results show that whether it is static datasets or neuromorphic datasets; whether it is direct training or ANN-SNN conversion training, our theory consistently demonstrated improvements in the experimental results. It is worth mentioning that even at ultra-low time steps, our method remains highly competitive compared to other methods.

\end{itemize}

\section{Related work}

\subsection{Training method of SNNs}
The training algorithms for SNNs can be categorized into unsupervised and supervised learning algorithms. Unsupervised learning algorithms are mostly bio-inspired, and the most commonly used one is the STDP (Spike-Timing-Dependent Plasticity) algorithm~\cite{song2000competitive} based on synaptic plasticity mechanisms~\cite{hebb2005organization}. 
These algorithms offer better biological interpretability, but their performance is generally less effective compared to supervised learning algorithms~\cite{liu2021sstdp}.
As for supervised learning, due to the non-differentiable nature of spikes, the supervised learning methods for SNNs differ significantly from those for ANNs. Currently, there are two training methods: one is direct training~\cite{wu2018spatio,yao2023attention,wang2022ltmd,zuo2020spiking,zuo2022multi}, which introduces a differentiable function to approximate the heaviside function. The derivative of this function is used to replace the derivative of the heaviside function at the corresponding position for backpropagation, hence it is also known as the gradient surrogate method. The other method is ANN-SNN conversion~\cite{bu2023optimal,liu2022spikeconverter,kim2020spiking}, which first trains an ANN, then replaces the network's activation function with spiking neurons, allowing the activation values of ANNs to be mapped to the firing rates of spiking neurons and continue to propagate in the network.

\subsection{Quantization loss}
Previous researchers~\cite{zheng2021going} have proposed and demonstrated that the membrane potential \(u\) follows a normal distribution with a mean of 0.
The occurrences of membrane potential exceeding the threshold are rare in the distribution, leading to a low firing rate of neurons as shown in Fig.\ref{3a}, which endows the spike with sparse characteristics. The sparse nature is the source of high performance and low power consumption in SNNs, but it also causes a significant quantization loss, hindering the development of deeper networks. To reduce quantization loss, many researchers have made efforts.

Some attempt to adjust the distribution of the membrane potential to alter the firing rate.~\cite{zheng2021going} indirectly adjust the distribution of membrane potential \(u\) by adjusting the variance of the input \(x\), ensuring that the firing rate remains at a reasonable level.
~\cite{guo2022loss} calculates the mutual information between the membrane potential and spikes to conclude that the maximum information content of spikes occurs when the firing rate is 0.5. 

Others try to expand spikes into multi-valued ones to enhance the information content of the spikes.~\cite{feng2022multi} packages k LIF neurons with different thresholds into a single MLF unit, where neurons within the unit share inputs and their outputs are summed, resulting in a spike value range of $\{0, 1, 2, \ldots, k\}$, which increases the diversity of the output spikes.
~\cite{zhao2022backeisnn} proposes an inhibitory type of spike, expanding the range of spike values to $\{-1,0,1\}$. Similarly, \cite{guo2024ternary} also introduces ternary spikes, using \(\alpha\) as a scaling coefficient to set the spike values to $\{-\alpha, 0, \alpha\}$. 

\begin{figure}[t]
  \centering
  \includegraphics[width=\textwidth]{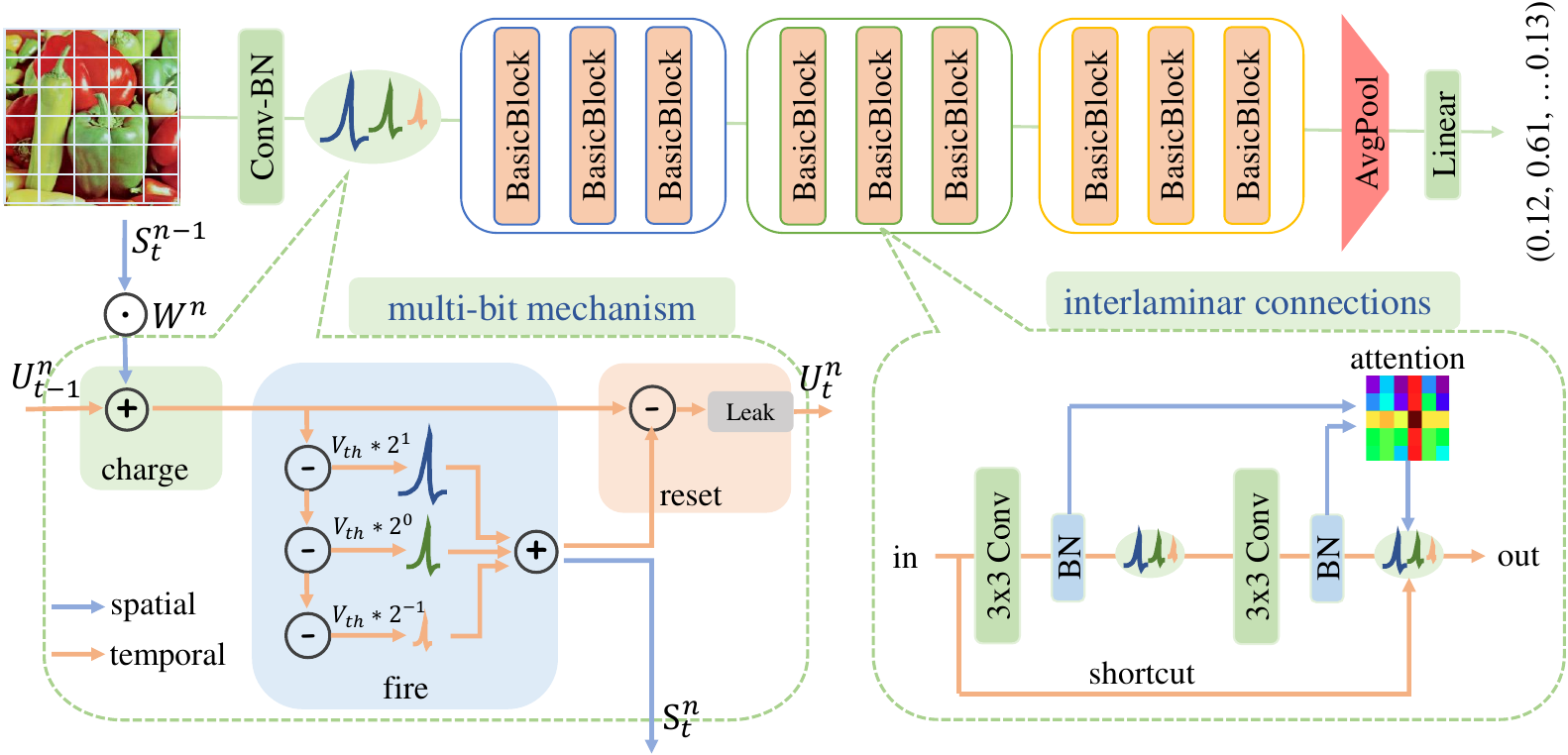}
  \caption{Overview of the proposed method. Our approach is grounded in the Spiking ResNet-20 architecture, where we enhance LIF neurons with a multi-bit mechanism to build an MBLIF model, thereby increasing the information density of neural spikes. Additionally, we incorporate interlaminar connections within the ResNet's basic block, aiming to preserve information maximally and ensure MBLIF's efficient spike generation across all bit positions.}
  \label{overview}
\end{figure}

\section{Method}

To reduce the loss during the forward process and enhance the amount of information carried by spikes, we propose a multi-bit information transmission mechanism. Futhermore, in order to preserve the information and promote the full firing of spikes across various bit positions, we introduce interlaminar connections. Specifically, as shown in Fig.\ref{overview}, we construct a ResNet-20 network, incorporating the multi-bit mechanism on the basis of LIF neurons, and introducing interlaminar connections  in the basic block of the ResNet.

\subsection{Multi-bit mechanism}
We propose multi-bit mechanism. This mechanism extends the output of neurons from single-bit to multi-bit. Specifically, due to the fact that connections between two neurons (synapses) are multivalent rather than unique, we extend the spike to a fixed-point unsigned binary number with \(m\) integer bits and \(n\) fractional bits. Specifically, the feature space has the shape \(C \times H \times W\), and the original spike is a single integer bit, with the representation range being \(S \in \{0, 1\}\), so the overall feature space distribution is \(X \in \{0, 1\}^{c \times h \times w}\). The extended spike consists of \(m\) integer bits and \(n\) fractional bits, with the representation range being \(S \in 2^{-n} \times \{0, 1, 2, \ldots, 2^{m+n}-1\} \), thus the overall feature space distribution is \(X \in 2^{-n} \times \{0, 1, 2, \ldots, 2^{m+n}-1\}^{c \times h \times w}\). Clearly, the representation capability of the extended spikes has significantly increased.

We are not merely increasing the number of integer bits \(m\) or fractional bits \(n\) independently to improve spike distinguishability. Instead, we combine both aspects because they serve different purposes. In the following, we will provide a detailed explanation of the differences between them.

\begin{figure}[h]
    \centering
    \begin{subfigure}[b]{0.32\textwidth}
        \includegraphics[width=\textwidth]{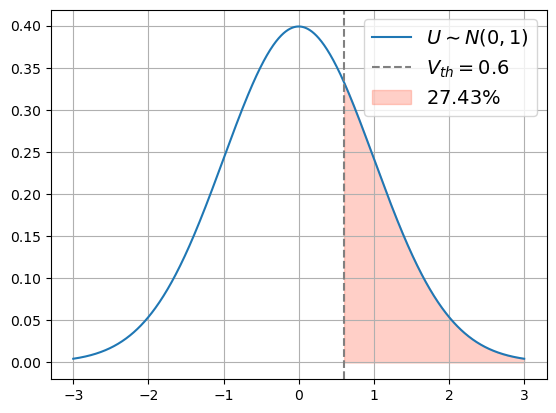}
        \caption{$S$}
        \label{3a}
    \end{subfigure}
    \hfill 
    \begin{subfigure}[b]{0.32\textwidth}
        \includegraphics[width=\textwidth]{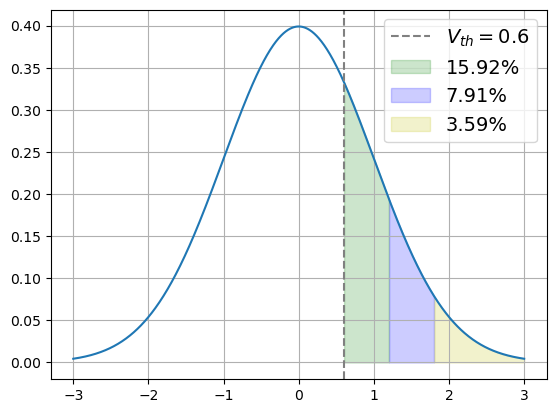}
        \caption{$S_{2\_0}$}
        \label{3b}
    \end{subfigure}
    \hfill 
    \begin{subfigure}[b]{0.32\textwidth}
        \includegraphics[width=\textwidth]{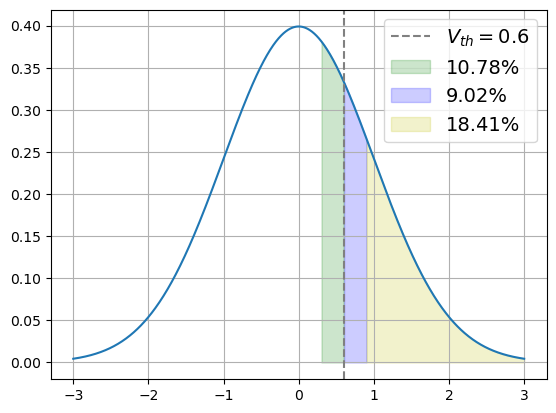}
        \caption{$S_{1\_1}$}
        \label{3c}
    \end{subfigure}
    \caption{(a)The range of values for the original spikes is $\{0,1\}$; (b)Expanding the spike by one integer bit upwards, the range of spike values is expanded to $\{0,1,2,3\}$; (c)Expanding the spike by one decimal place downward, the range of spike values expands to $\{0,0.5,1,1.5\}$.}
    \label{fig:three_images}
\end{figure}

We first review the firing scenario of the basic LIF model as presented in Fig.\ref{3a}, where we assume without loss of generality that the membrane potential follows a standard normal distribution, with $V_{th}=0.6$. We use entropy to represent the loss of information before and after firing. Before firing, the membrane potential is normally distributed. Its probability density function is 
\begin{eqnarray}
    f(u) = \frac{1}{\sqrt{2\pi}} e^{-\frac{u^2}{2}}
\end{eqnarray}
and its entropy is fixed, which can be described as
\begin{eqnarray}
    H(U) = -\int_{-\infty}^{\infty} f(u) \log f(u) \, du = \frac{1}{2} \log_2(2\pi e)
\end{eqnarray}
In the LIF model, the continuous $u$ is discretized into spikes, whose probabilities can be expressed as
\begin{eqnarray}
    \begin{cases}
        P_1 = P(S=1) = P(U \ge V_{th}) = 1 - \Phi(V_{th}) \\
        P_0 = P(S=0) = P(U < V_{th}) = \Phi(V_{th})
    \end{cases}
\end{eqnarray}
where \(\Phi\) denotes the upper quantile of the normal distribution. The entropy of spikes is
\begin{eqnarray}
    H(S) = -\sum_{i = 0,1}P_i log_2 P_i = - P_0 \log_2 P_0 - P_1 \log_2 P_1
\end{eqnarray}
The forward process loss of LIF is
\begin{eqnarray}
    Loss &= H(U) - H(S) = \frac{1}{2} \log(2\pi e) +  P_0 \log_2 P_0 + P_1 \log_2 P_1
\end{eqnarray}
The magnitude of Loss is solely determined by \(H(S)\), as \(H(U)\) is fixed. Therefore, increasing the magnitude of \(H(S)\) is beneficial for reducing the loss in the forward process.


Next, let's analyze the case of expanding the number of integer bits. Assuming an upward expansion of one bit, i.e., m=2, the range of spike values becomes \{0, 1, 2, 3\}. Thus, Eq.\ref{lif:fire} is transformed to
\begin{eqnarray}
    S_{2\_0}=
        \begin{cases}
            3&  \frac{u}{V_{th}} \geq 3, \\
            2&  2\leq \frac{u}{V_{th}} <3, \\
            1&  1\leq \frac{u}{V_{th}} <2, \\
            0&  \frac{u}{V_{th}} < 1. 
        \end{cases}
    \label{lif_2_0}
\end{eqnarray}
where \(S_{2\_0}\) represents a spike containing 2-bit integer bits, the maximum value that can be encoded is 3, thus when the membrane potential reaches three times the threshold, the maximum value of the spike is 3. The specific divisions are  illustrated in Fig.\ref{3b}. The entropy can be expressed as
\begin{eqnarray}
    H(S_{2\_0}) = -\sum_{i = 0,1,2,3}P_i log_2 P_i
\end{eqnarray}
Expanding the number of integer bits does not change the quantity of spikes. Rather, it granularizes the existing spikes. This method significantly enhances the distinction of spikes, increasing the amount of information carried by them and reducing the loss in the conversion from membrane potential to spikes. However, the potential for expansion in this manner is limited, as only a very few neurons are capable of accumulating a membrane potential that exceeds several times the threshold.

Next, we analyze the expansion of spikes in terms of decimal places, as shown in Fig.\ref{3c}, which illustrates the result of expanding by one decimal place. The implementation method is similar to that of expanding the integer bits, but the effects are different. Expanding the decimal places not only allows for further granularization of the existing spikes but also enables some neurons that have not reached the threshold, yet are relatively active, to transmit part of the signal downstream. The corresponding forward process is
\begin{eqnarray}
    S_{1\_1}=
        \begin{cases}
            1.5&  \frac{u}{V_{th}} \geq 1.5, \\
            1&  0.5\leq \frac{u}{V_{th}} <1, \\
            0.5&  0.5\leq \frac{u}{V_{th}} <1, \\
            0&  \frac{u}{V_{th}} < 0.5. 
        \end{cases}
    \label{lif_1_1}
\end{eqnarray}
and the entropy can be expressed as
\begin{eqnarray}
    H(S_{1\_1}) = -\sum_{i = 0,0.5,1,1.5}P_i log_2 P_i
\end{eqnarray}

\begin{figure}[h]
    \centering
    \includegraphics[width=\columnwidth]{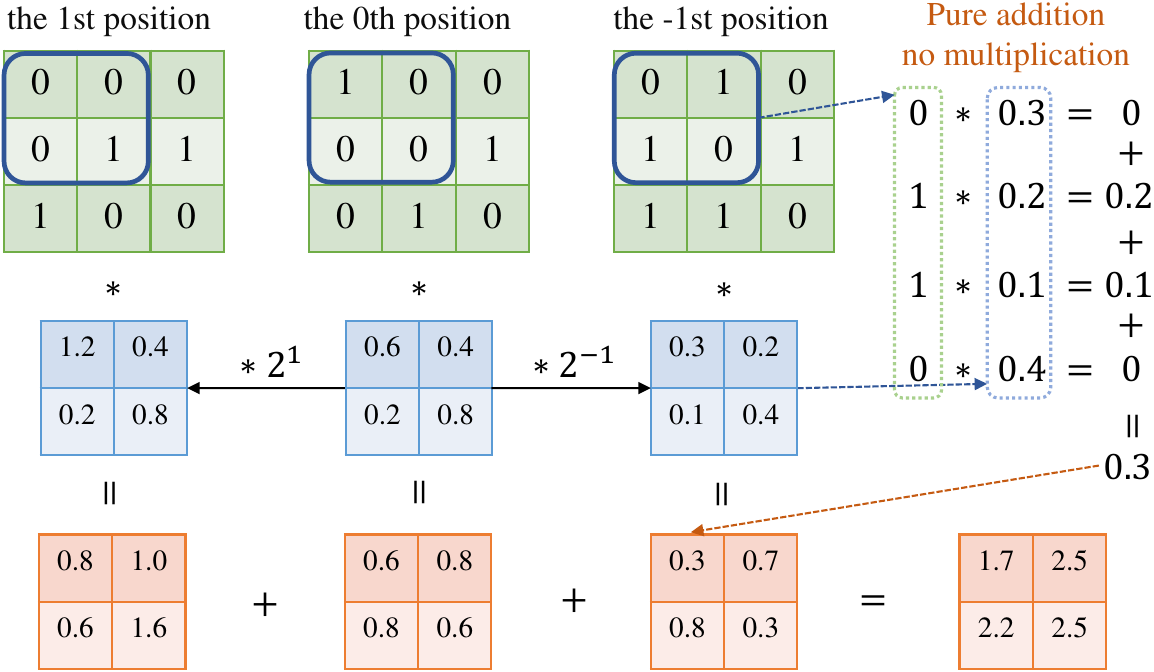}
    \caption{The diagram represents the case of \( m=2, n=1 \), where the green matrix denotes the activation values corresponding to three bit positions with weights \( 2^1, 2^0, \) and \( 2^{-1} \). These weights are absorbed by the blue convolutional layer parameter matrix and remain unchanged during the inference process. Thus, the final computation between the multi-bit spikes and the convolutional layer parameters is still addition.}
    \label{cal}
\end{figure}
It is worth noting that the additional computational cost brought by the multi-bit mechanism is very minimal, as it does not alter the advantage of SNNs in converting MAC operations to AC operations. With the introduction of the multi-bit mechanism, the binary weights of each bit can be perfectly integrated into the parameters. SNNs still use addition for computations, as demonstrated in the Fig.\ref{cal} showing the calculation process of convolutional layer parameters and multi-bit spikes.

Finally, we can calculate the entropy corresponding to the above scenarios. Clearly, as shown in Tab.\ref{entropy}, with the bit width increases, the entropy of the spikes is increasing, which means that the information loss in converting membrane potential to spikes is decreasing.

\begin{table}[h] 
\centering 
\caption{Entropy variation with spike bit width. As the bit width increases, spike entropy rises, indicating an increase in information.}
\begin{tabular}{cccc}
\toprule
\(H(S)\) & \(H(S_{2\_0})\) & \(H(S_{1\_1})\) & \(H(S_{2\_1})\) \\
\midrule
0.848 & 1.149 & 1.361 & 1.998 \\
\bottomrule
\end{tabular}

\label{entropy}
\end{table}

\subsection{Interlaminar connections}
Whether it's an ANN or an SNN, information in their network structures typically flows layer by layer. However, in real neural networks, cross-layer information fusion truly exists, known as interlaminar connections~\cite{thomson2003interlaminar}. Here, to further reduce the information loss in the forward process and facilitate the potential of multi-bit mechanisms, we incorporate interlaminar connections into the Basic Block of ResNet. Specifically, the original structure of the Basic Block remains unchanged, but the outputs from the two BN layers are merged before being fed into the second spiking neuron. The fusion process is illustrated in Algorithm.~\ref{algorithm_ic}, where outputs from the two BN layers are first concatenated along the channel dimension, followed by a 1D convolution and BN. Next, the ECA module is applied to distinguish salient features while ignoring less important ones, obtaining the final fused result, which is then input into MBLIF to complete the fusion of information. It should be noted that in the fusion process, although some additional modules are introduced, apart from the 1D convolution and ECA module which have a small number of learnable parameters, the overall increase in parameters and computational cost is negligible . The entire process can be expressed as
\begin{eqnarray}
    X_{t} = BN(Conv_{1*1}(ConCat(X_{t}^{pre},X_{t}^{post}))),
\end{eqnarray}
\begin{eqnarray}
     X_{t}' =  X_{t} \odot g(X_{t})
\end{eqnarray}
Wherein, $X_{t}^{pre}$ and $X_{t}^{post}$ respectively represent the outputs of the preceding and following BN layers, and $g(\cdot)$ denotes the ECA attention process as shown in Eq.~\ref{eq:eca}.



\begin{algorithm}[H]
\label{algorithm_ic}
\caption{Forward Function of the Interlaminar Connections}

\SetKwInOut{Input}{Input}
\SetKwInOut{Output}{Output}

\Input{$x$ - Input spike of the interlaminar connections module.}
\Output{$out$ - Output spike of the interlaminar connections module.}

\SetKwFunction{Conv1}{$Conv_{3*3}$}
\SetKwFunction{Conv2}{conv2}
\SetKwFunction{Conv}{conv}
\SetKwFunction{BN1}{bn1}
\SetKwFunction{BN2}{bn2}
\SetKwFunction{BN}{bn}
\SetKwFunction{MBLIF}{MBLIF}
\SetKwFunction{ECA}{ECA}
\SetKwFunction{Shortcut}{shortcut}
\SetKwFunction{Concat}{Concat}

\SetKwProg{Fn}{def}{:}{}
\Fn{forward($x$)}{
    $x^{\text{pre}} \gets \text{BN}(\text{Conv}_{3 \times 3}(x))$; \# The input undergoes the first convolution to generate $x^{\text{pre}}$.
    $x \gets \MBLIF(x^{\text{pre}})$\;
    $x^{\text{post}} \gets \text{BN}(\text{Conv}_{3 \times 3}(x))$; \# The input undergoes the second convolution to generate $x^{\text{pre}}$. 
    $x^{\text{re}} \gets \text{BN}(\text{Conv}_{1 \times 1}(\Concat([x^{\text{pre}}, x^{\text{post}}], \text{dim}=1)))$; \# Concatenate $x^{\text{pre}} $and $x^{\text{post}} $. 
    $x^{\text{re}} \gets x^{\text{re}} \odot Eq.~\ref{eq:eca}(x^{\text{re}})$; \# Extract effective information from the fused data using ECA.
    $x^{\text{shortcut}} \gets \Shortcut(x)$\; 
    $x \gets x^{\text{post}} + x^{\text{shortcut}} + x^{\text{re}}$; \# Accumulate the information and then input it into MBLIF. 
    $out \gets \MBLIF(x)$\; 
    \Return $out$\;
}
\end{algorithm}


\begin{figure}[h]
    \centering
    \resizebox{\textwidth}{!}{  
        \begin{minipage}{\textwidth}
            \centering
                \begin{subfigure}[b]{0.25\textwidth}
                    \includegraphics[width=\textwidth]{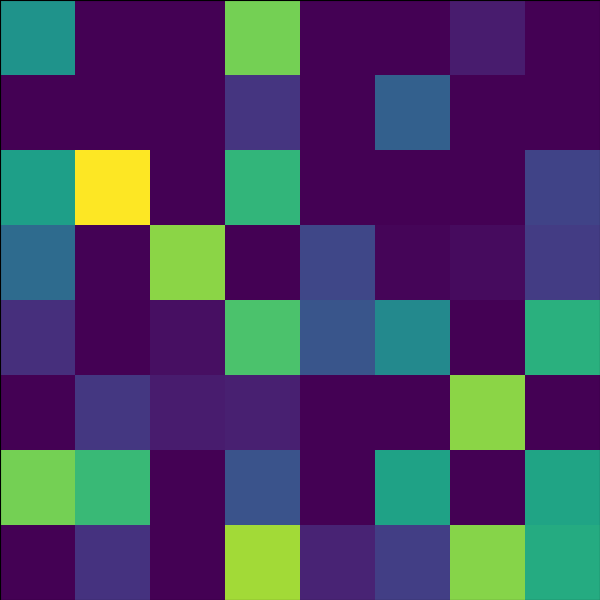}
                    \caption{Before charging}
                    \label{heatmap1}
                \end{subfigure}
                \hfill 
                \begin{subfigure}[b]{0.25\textwidth}
                    \includegraphics[width=\textwidth]{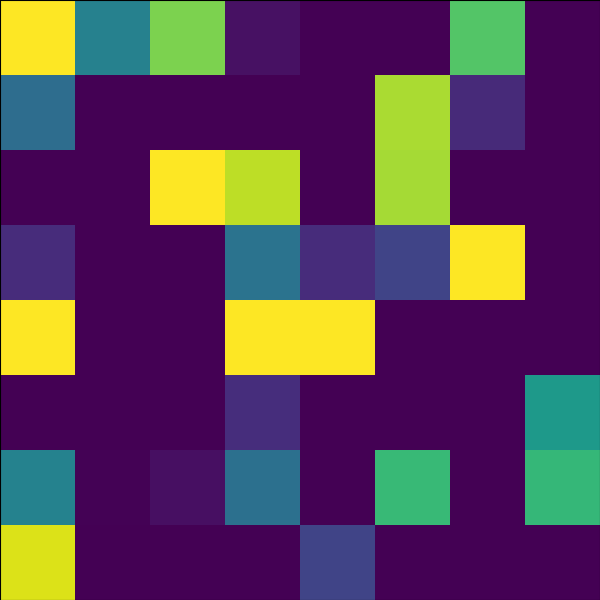}
                    \caption{After charging}
                    \label{heatmap2}
                \end{subfigure}
                \hfill 
                \begin{subfigure}[b]{0.25\textwidth}
                    \includegraphics[width=\textwidth]{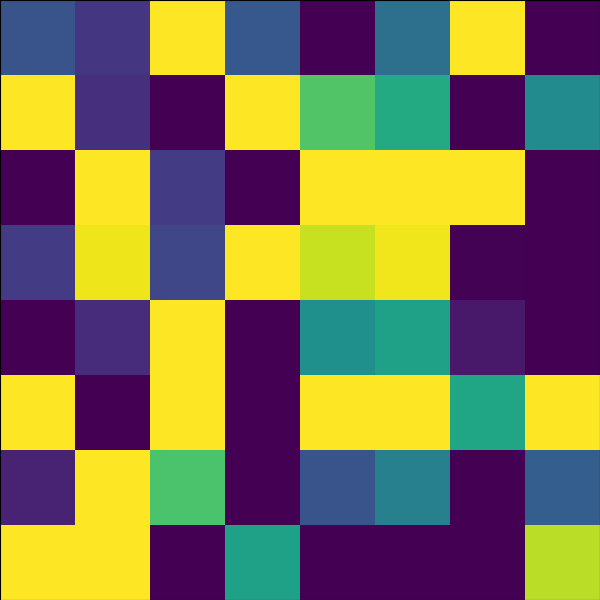}
                    \caption{Re-stimulate}
                    \label{heatmap3}
                \end{subfigure}
        \end{minipage}
    }
    \caption{Heatmaps of the membrane potentials in a certain layer, with colors ranging from blue to yellow indicating neurons transitioning from silent to active.}
    \label{fig:three_heatmap}
\end{figure}

Interlaminar connections not only directly enrich the network's topological connections and enhance its expressive capabilities, but also indirectly alter the dynamical activity of spiking neurons. Specifically, on top of existing activity, newly introduced connections utilize the attention mechanism to extract effective features, which are then used to re-stimulate the neurons. As shown in Fig.\ref{heatmap1}-Fig.\ref{heatmap3}, this stimulation serves as a compensation for lost information and essentially changes the original charging method of the neurons, acting as a kind of secondary charging. This means that neurons will be more fully activated; already active neurons will become even more active, and neurons that were previously inactive may also enter an active state. This encourages neurons to fully fire across all bits, fully exploiting the potential of the multi-bit information transmission mechanism.


\section{Experiments}
In this section, we conducted extensive experiments. Whether on static datasets or neuromorphic datasets, and whether using direct training methods or ANN-SNN conversion training, the results consistently showed improvements. Details are provided in the appendix.

\subsection{Static image classification}
\begin{table}[h]
    \centering
    \caption{Performance on the CIFAR-10(left) and CIFAR-100(right) with direct training method.}
    \begin{minipage}{.48\textwidth}
        \centering

        





        \begin{tabular}{cccc}
        \toprule
         Work             & Network   & T    & Accuracy\\
        
        \midrule  
        TSSL-BP~\cite{zhang2020temporal}        & CIFARNet    & 5   & 91.41$\%$ \\
        MLF~\cite{feng2022multi}        & ResNet-20    & 4   & 94.25$\%$ \\  \cline{3-4}

        \multirow{3}{*}{tdBN~\cite{zheng2021going}}   & \multirow{3}{*}{ResNet-19}   & 2   & 92.34$\%$ \\
        &                   & 4   & 92.92$\%$ \\
        &                   & 6   & 93.16$\%$ \\ \cline{3-4}

        \multirow{2}{*}{TET~\cite{deng2022temporal}}   & \multirow{2}{*}{ResNet-19}   & 4   & 94.44$\%$ \\
        &                   & 6   & 94.50$\%$ \\  \cline{3-4}

        \multirow{2}{*}{Ternary~\cite{guo2024ternary}}   & \multirow{2}{*}{ResNet-20}   & 2   & 94.48$\%$ \\
        &                   & 4   & 94.96$\%$ \\

        \midrule 
        \multirow{4}{*}{OURS}   & \multirow{4}{*}{ResNet-20}        & 1    & 94.59$\%$ \\
        &          & 2    & 94.75$\%$ \\
        &          & 4    & 94.93$\%$ \\
        &          & 6    & 95.00$\%$ \\                   
        \bottomrule
        \end{tabular}

    \end{minipage}%
    \hfill
        \begin{minipage}{.48\textwidth}
        \centering

    
        


                            

        \begin{tabular}{cccc}
        \toprule
         Work         & Network   & T    & Accuracy\\
        \midrule  
        IM-LOSS~\cite{guo2022loss}        & VGG-16    & 5   & 70.18$\%$ \\
        PALIF~\cite{ding2023improved}     & ResNet-20    & 4   & 76.09$\%$ \\  \cline{3-4}
    
        \multirow{3}{*}{tdBN~\cite{zheng2021going}}   & \multirow{3}{*}{ResNet-19}   & 2   & 69.41$\%$ \\
        &              & 4   & 70.86$\%$ \\
        &              & 6   & 71.12$\%$ \\  \cline{3-4}
        
        \multirow{3}{*}{TET~\cite{deng2022temporal}}  & \multirow{3}{*}{ResNet-19}   & 2   & 72.87$\%$ \\
        &              & 4   & 74.47$\%$ \\
        &              & 6   & 74.72$\%$ \\   \cline{3-4}

        \multirow{2}{*}{Ternary~\cite{guo2024ternary}}   & \multirow{2}{*}{ResNet-20}   & 2   & 73.41$\%$ \\
        &                   & 4   & 74.02$\%$ \\

        \midrule
        \multirow{3}{*}{OURS} & \multirow{3}{*}{ResNet-20}        & 1    & 75.43$\%$ \\
        &     & 2    & 76.06$\%$ \\
        &     & 4    & 76.51$\%$ \\
                            
        \bottomrule
        \end{tabular}

    \end{minipage}%
    \label{CIFAR-10-100}
\end{table}

We conducted classification experiments on static data using the direct training method. Tab.\ref{CIFAR-10-100} show our experimental results on CIFAR-10\cite{krizhevsky2009learning} and CIFAR-100\cite{krizhevsky2009learning}, which indicate that we achieve 95.00$\%$ and 76.51$\%$ accuracy on these datasets, respectively.

\subsection{Neuromorphic data classification}

\begin{figure}[h]
    \centering
    \includegraphics[width=0.8\columnwidth]{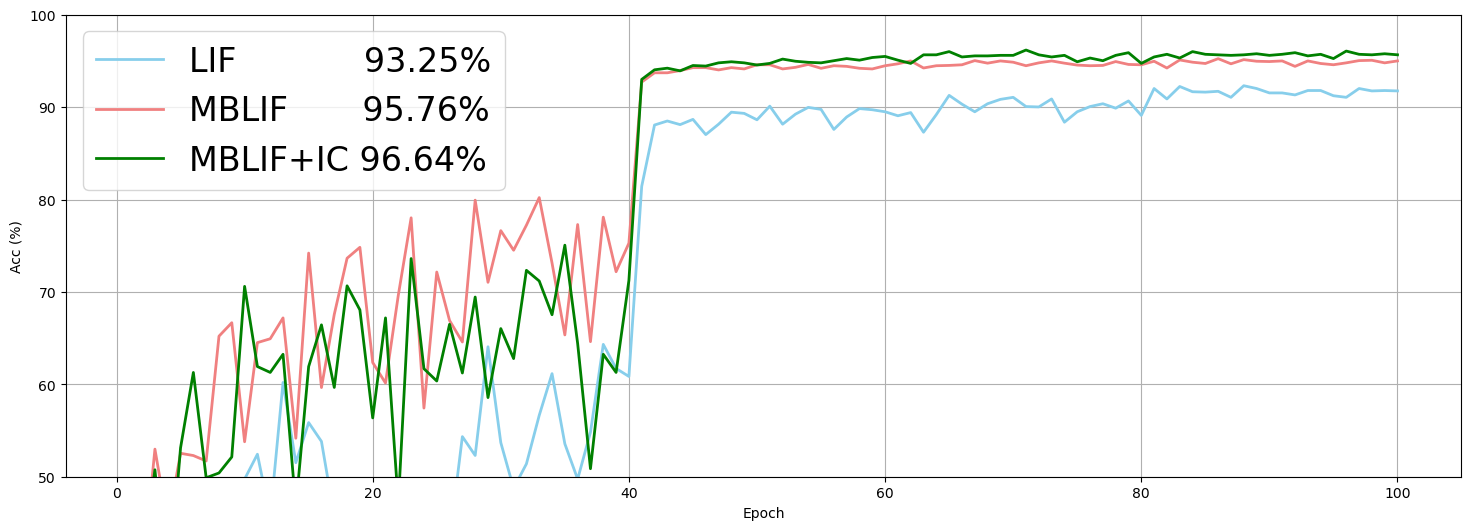}
    \caption{Performance on the DVS-Gesture dataset. Our method, particularly the multi-bit mechanism, can significantly enhance the network's ability to recognize neuromorphic data.}
    \label{gesture}
\end{figure}

We conduct experiments on the DVS-Gesture dataset\cite{Amir2017lowpower} using the direct training method. This is a more information-rich neuromorphic dataset for which we use the method of~\cite{feng2022multi} to preprocess data. 
The experimental results are shown in Fig.\ref{gesture}. Our multi-bit mechanism results in a 2.51$\%$ increase in classification accuracy, and the interlaminar connections further increase the accuracy by 0.88$\%$. This demonstrates that our method is also applicable to neuromorphic datasets.

\subsection{Ablation experiment}
We conduct ablation experiments. We base our experiments on the ResNet-20 network and LIF neurons, with all parameters uniformly set: $V_{th}=0.6,\tau=4$ and $t=4$. We particularly focus on ablating the bit width of spikes, with experimental details presented in Tab.\ref{Ablation}. It can be observed that the baseline LIF (1) results are 93.21$\%$/72.71$\%$. The expansion of bit width greatly improves the classification accuracy, with an upward expansion of one integer bit (3) leading to an increase of 1.13$\%$/2.59$\%$, a downward expansion of one fractional bit (4) resulting in a 1.23$\%$/1.77$\%$ improvement, and simultaneous expansion in both directions (5) yielding an increase of 1.40$\%$/2.94$\%$. The introduction of interlaminar connections in LIF (2) leads to an improvement of 0.48$\%$/0.62$\%$ over the baseline (1). When all settings are introduced (6), the network achieves an improvement of 1.72$\%$ on CIFAR-10 and an even more significant increase of 3.80$\%$ on CIFAR-100.
\begin{table}[h]
    \centering
    \caption{Ablation experiment on CIFAR-10 and CIFAR-100. With the increase in bit width and the introduction of IC (interlaminar connections), the network's performance gradually improves.}
    \begin{tabular}{ccccccc}
        \toprule
        index & Neuron  & Bits & IC     & CIFAR-10     & CIFAR-100  \\
        
        \midrule
        1&LIF         & 1 + 0      & \ding{55}      & 93.21$\%$   & 72.71$\%$  \\
        2&LIF         & 1 + 0      & \checkmark     & 93.69$\%$   & 73.33$\%$  \\
        3&MBLIF       & 2 + 0      & \ding{55}      & 94.34$\%$   & 75.30$\%$  \\
        4&MBLIF       & 1 + 1      & \ding{55}      & 94.44$\%$   & 74.48$\%$  \\
        5&MBLIF       & 2 + 1      & \ding{55}      & 94.61$\%$   & 75.65$\%$  \\
        6&MBLIF       & 2 + 1      & \checkmark     & 94.93$\%$   & 76.51$\%$  \\
        
        \bottomrule
    \end{tabular}
    \label{Ablation}
\end{table}

\subsection{Performance at ultra-low time steps}
Notably, as shown in Tab.\ref{table:ultra-low}, even with just a single time step, our method remains highly competitive compared to many other works. This is primarily due to our multi-bit mechanism, which extends the bit width of spikes, trading space for time. The information contained in spikes at a single time step is significantly increased, enabling the SNN to achieve high accuracy with ultra-low time step.
\begin{table}[h!]
    \centering
    \caption{93.16$\%$ / 6 indicates an accuracy of 93.16$\%$ at a time step of 6. Compared to other works, our method remains competitive even with a single time step.}
    \begin{tabular}{lcccc}
        \toprule
        & tdBN~\cite{zheng2021going} & TET~\cite{deng2022temporal} & Ternary~\cite{guo2024ternary} & OURS \\
        \midrule
        CIFAR-10  & 93.16$\% / 6 $  & 94.50$\%$ / 6 & 94.48$\%$ / 2 & 94.59$\%$ / 1 \\
        CIFAR-100 & 71.12$\% / 6 $ & 74.72$\%$ / 6 & 74.02$\%$ / 4 & 75.43$\%$ / 1 \\
        \bottomrule
    \end{tabular}

    \label{table:ultra-low}
\end{table}

\subsection{Visualization}
\begin{figure}[h]
  \centering
    \resizebox{\textwidth}{!}{  
  
  \begin{subfigure}{.24\textwidth}
    \includegraphics[width=\linewidth]{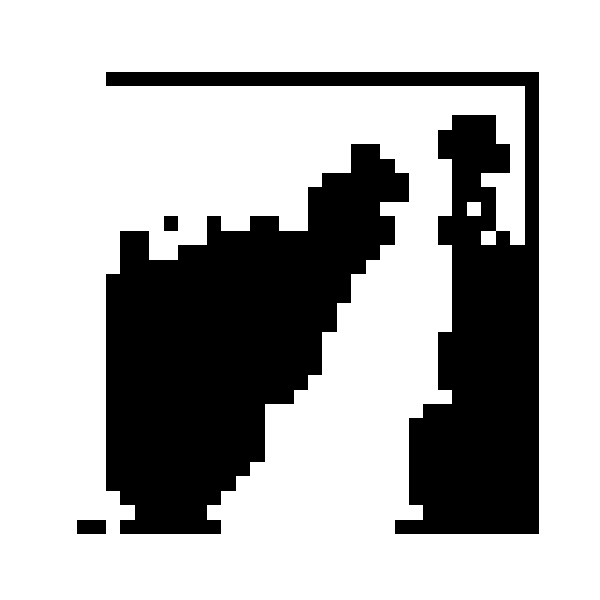}
    \caption{$H_{1\_0}=0.999$}
    \label{6_c}
  \end{subfigure}
  \begin{subfigure}{.24\textwidth}
    \includegraphics[width=\linewidth]{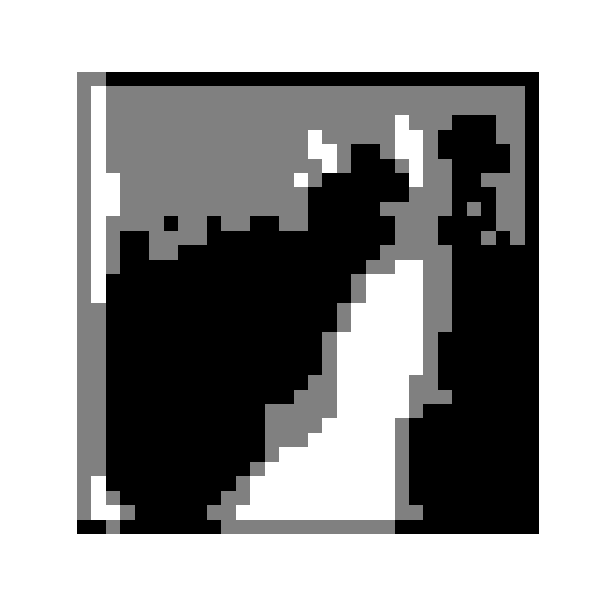}
    \caption{$H_{2\_0}=1.424$}
    \label{6_d}
  \end{subfigure}
  \begin{subfigure}{.24\textwidth}
    \includegraphics[width=\linewidth]{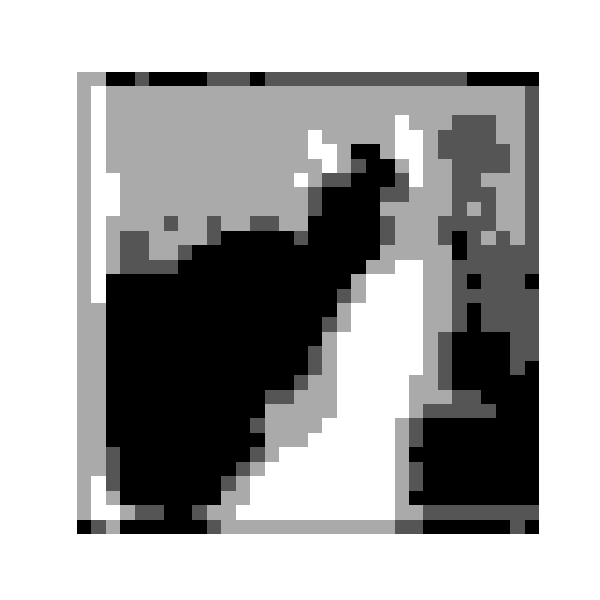}
    \caption{$H_{1\_1}=1.884$}
    \label{6_e}
  \end{subfigure}
  \begin{subfigure}{.24\textwidth}
    \includegraphics[width=\linewidth]{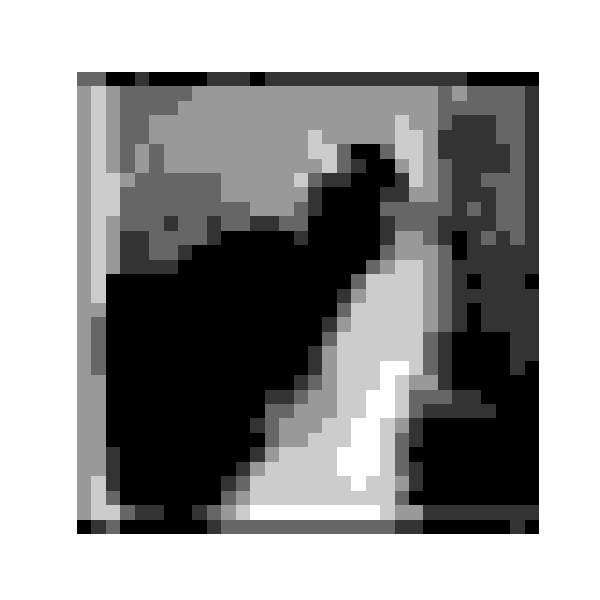}
    \caption{$H_{2\_1}=2.324$}
    \label{6_f}
  \end{subfigure}
  }

  \caption{Visualization of spike firing at different bit widths. As the bit width increases, the entropy of the spikes increases, and the details in the visualized images become progressively richer.}
  \label{fig_6}
\end{figure}
We attempt to visualize the spike firing under the multi-bit mechanism. We extract the membrane potentials of neurons in a certain layer of the network after receiving current inputs and vary the bit width of the spikes to observe the firing activity.
As the bit width increases from Fig.\ref{6_c} to Fig.\ref{6_f}, the entropy of the spikes significantly increases, and the details presented in the grayscale images become increasingly rich. After expanding by two bits, the entropy in Fig.\ref{6_f} rises by 132$\%$ compared to the baseline in Fig.\ref{6_c}, indicating a substantial increase in the information content of the spikes.


\subsection{Compared to similar ideas}
As the Tab.\ref{multi-value} shows, we compare the proposed method with some similar ideas, which mostly expand the spike, increasing its representation range. Although the basic LIF operates faster in AC operations, its information capacity is small, only representing {0,1}. The extension by \cite{feng2022multi}, while enhancing representational capability, reverts AC operations back to the original MAC operations, and the dynamics of $k$ neurons obviously increase the computational load, which is clearly not deployment-friendly. The ideas of \cite{zhao2022backeisnn} and \cite{guo2024ternary} are similar, both expanding the spike into ternary, which does not introduce much extra computation and retains the advantages of AC, but their range is limited to ternary with no further expansion space. Our multi-bit information transfer mechanism, while preserving the advantages of AC operations, also benefits from the binary nature, having nearly no increase in computation compared to the basic LIF and possesses good scalability with a wide range of feature representation capabilities.
\begin{table}[ht]
    \centering
    \caption{Compared to other multi-valued output methods, our multi-bit mechanism has advantages in both computational simplification and representational capacity.}
    \begin{tabular}{cccc}
        \toprule
        Work    & Operation        & Capacity  &Explanation \\
        \midrule
        BASIC       &AC        &$\{0, 1\}$   & original spike\\
        MLF~\cite{feng2022multi}       &MAC    & $\{0, 1, 2, \ldots, k\}$  &sum of of k LIF neurons \\
        BackEI~\cite{zhao2022backeisnn}    &AC  &$\{-1,0,1\}$    &ternary spike \\
        Ternary~\cite{guo2024ternary}   &AC  &$\alpha \times \{-1,0,1\}$  &scalable ternary spike \\
        OURS           &AC     &$2^{-n} \times \{0, 1, 2, \ldots, 2^{m+n}-1\}$  &m-bit integer and n-bit fraction\\
        \bottomrule
    \end{tabular}
    \label{multi-value}
\end{table}

\section{Conclusion}
We propose a multi-bit information transmission mechanism that expands spikes from a single bit to multiple bits, greatly enriching the information content of the spikes. We also introduce interlaminar connections to facilitate the full release of spikes across various bit positions. To our knowledge, this is the first work that attempts to expand the bit-width of spikes, representing a novel information transmission paradigm for SNNs. Compared to previous models, our method introduces very few parameters yet significantly reduces the quantization loss in the forward process and still performs excellently at ultra-low time steps. 

However, the increase in bit-width inevitably leads to a slight increase in spike memory usage and computation time during inference. Nevertheless, considering the reduction in information loss and the overall performance improvement, this should be tolerable. We look forward to further improvements in the future and hope that our work can inspire subsequent information transmission mechanisms, contributing to the research and development of high-precision, low-latency SNNs.


\bibliographystyle{unsrt}
\bibliography{multibits}

\begin{thebibliography}{10}

\bibitem{krizhevsky2012imagenet}
Alex Krizhevsky, Ilya Sutskever, and Geoffrey~E Hinton.
\newblock Imagenet classification with deep convolutional neural networks.
\newblock {\em Advances in neural information processing systems}, 25, 2012.

\bibitem{silver2016mastering}
David Silver, Aja Huang, Chris~J Maddison, Arthur Guez, Laurent Sifre, George
  Van Den~Driessche, Julian Schrittwieser, Ioannis Antonoglou, Veda
  Panneershelvam, Marc Lanctot, et~al.
\newblock Mastering the game of go with deep neural networks and tree search.
\newblock {\em nature}, 529(7587):484--489, 2016.

\bibitem{maass1997networks}
Wolfgang Maass.
\newblock Networks of spiking neurons: the third generation of neural network
  models.
\newblock {\em Neural networks}, 10(9):1659--1671, 1997.

\bibitem{roy2019towards}
Kaushik Roy, Akhilesh Jaiswal, and Priyadarshini Panda.
\newblock Towards spike-based machine intelligence with neuromorphic computing.
\newblock {\em Nature}, 575(7784):607--617, 2019.

\bibitem{diehl2015unsupervised}
Peter~U Diehl and Matthew Cook.
\newblock Unsupervised learning of digit recognition using
  spike-timing-dependent plasticity.
\newblock {\em Frontiers in computational neuroscience}, 9:99, 2015.

\bibitem{thomson2003interlaminar}
Alex~M Thomson and A~Peter Bannister.
\newblock Interlaminar connections in the neocortex.
\newblock {\em Cerebral cortex}, 13(1):5--14, 2003.

\bibitem{wang2020eca}
Qilong Wang, Banggu Wu, Pengfei Zhu, Peihua Li, Wangmeng Zuo, and Qinghua Hu.
\newblock Eca-net: Efficient channel attention for deep convolutional neural
  networks.
\newblock In {\em Proceedings of the IEEE/CVF conference on computer vision and
  pattern recognition}, pages 11534--11542, 2020.

\bibitem{song2000competitive}
Sen Song, Kenneth~D Miller, and Larry~F Abbott.
\newblock Competitive hebbian learning through spike-timing-dependent synaptic
  plasticity.
\newblock {\em Nature neuroscience}, 3(9):919--926, 2000.

\bibitem{hebb2005organization}
Donald~Olding Hebb.
\newblock {\em The organization of behavior: A neuropsychological theory}.
\newblock Psychology press, 2005.

\bibitem{liu2021sstdp}
Fangxin Liu, Wenbo Zhao, Yongbiao Chen, Zongwu Wang, Tao Yang, and Li~Jiang.
\newblock Sstdp: Supervised spike timing dependent plasticity for efficient
  spiking neural network training.
\newblock {\em Frontiers in Neuroscience}, 15:756876, 2021.

\bibitem{wu2018spatio}
Yujie Wu, Lei Deng, Guoqi Li, Jun Zhu, and Luping Shi.
\newblock Spatio-temporal backpropagation for training high-performance spiking
  neural networks.
\newblock {\em Frontiers in neuroscience}, 12:331, 2018.

\bibitem{yao2023attention}
Man Yao, Guangshe Zhao, Hengyu Zhang, Yifan Hu, Lei Deng, Yonghong Tian, Bo~Xu,
  and Guoqi Li.
\newblock Attention spiking neural networks.
\newblock {\em IEEE transactions on pattern analysis and machine intelligence},
  2023.

\bibitem{wang2022ltmd}
Siqi Wang, Tee~Hiang Cheng, and Meng-Hiot Lim.
\newblock Ltmd: Learning improvement of spiking neural networks with learnable
  thresholding neurons and moderate dropout.
\newblock {\em Advances in Neural Information Processing Systems},
  35:28350--28362, 2022.

\bibitem{zuo2020spiking}
Lin Zuo, Yi~Chen, Lei Zhang, and Changle Chen.
\newblock A spiking neural network with probability information transmission.
\newblock {\em Neurocomputing}, 408:1--12, 2020.

\bibitem{zuo2022multi}
Lin Zuo, Fengjie Xu, Changhua Zhang, Tangfan Xiahou, and Yu~Liu.
\newblock A multi-layer spiking neural network-based approach to bearing fault
  diagnosis.
\newblock {\em Reliability Engineering \& System Safety}, 225:108561, 2022.

\bibitem{bu2023optimal}
Tong Bu, Wei Fang, Jianhao Ding, PengLin Dai, Zhaofei Yu, and Tiejun Huang.
\newblock Optimal ann-snn conversion for high-accuracy and ultra-low-latency
  spiking neural networks.
\newblock {\em arXiv preprint arXiv:2303.04347}, 2023.

\bibitem{liu2022spikeconverter}
Fangxin Liu, Wenbo Zhao, Yongbiao Chen, Zongwu Wang, and Li~Jiang.
\newblock Spikeconverter: An efficient conversion framework zipping the gap
  between artificial neural networks and spiking neural networks.
\newblock In {\em Proceedings of the AAAI Conference on Artificial
  Intelligence}, volume~36, pages 1692--1701, 2022.

\bibitem{kim2020spiking}
Seijoon Kim, Seongsik Park, Byunggook Na, and Sungroh Yoon.
\newblock Spiking-yolo: spiking neural network for energy-efficient object
  detection.
\newblock In {\em Proceedings of the AAAI conference on artificial
  intelligence}, volume~34, pages 11270--11277, 2020.

\bibitem{zheng2021going}
Hanle Zheng, Yujie Wu, Lei Deng, Yifan Hu, and Guoqi Li.
\newblock Going deeper with directly-trained larger spiking neural networks.
\newblock In {\em Proceedings of the AAAI conference on artificial
  intelligence}, volume~35, pages 11062--11070, 2021.

\bibitem{guo2022loss}
Yufei Guo, Yuanpei Chen, Liwen Zhang, Xiaode Liu, Yinglei Wang, Xuhui Huang,
  and Zhe Ma.
\newblock Im-loss: information maximization loss for spiking neural networks.
\newblock {\em Advances in Neural Information Processing Systems}, 35:156--166,
  2022.

\bibitem{feng2022multi}
Lang Feng, Qianhui Liu, Huajin Tang, De~Ma, and Gang Pan.
\newblock Multi-level firing with spiking ds-resnet: Enabling better and deeper
  directly-trained spiking neural networks.
\newblock In Lud~De Raedt, editor, {\em Proceedings of the Thirty-First
  International Joint Conference on Artificial Intelligence, {IJCAI-22}}, pages
  2471--2477. International Joint Conferences on Artificial Intelligence
  Organization, 7 2022.
\newblock Main Track.

\bibitem{zhao2022backeisnn}
Dongcheng Zhao, Yi~Zeng, and Yang Li.
\newblock Backeisnn: A deep spiking neural network with adaptive self-feedback
  and balanced excitatory--inhibitory neurons.
\newblock {\em Neural Networks}, 154:68--77, 2022.

\bibitem{guo2024ternary}
Yufei Guo, Yuanpei Chen, Xiaode Liu, Weihang Peng, Yuhan Zhang, Xuhui Huang,
  and Zhe Ma.
\newblock Ternary spike: Learning ternary spikes for spiking neural networks.
\newblock In {\em Proceedings of the AAAI Conference on Artificial
  Intelligence}, volume~38, pages 12244--12252, 2024.

\bibitem{zhang2020temporal}
Wenrui Zhang and Peng Li.
\newblock Temporal spike sequence learning via backpropagation for deep spiking
  neural networks.
\newblock {\em Advances in Neural Information Processing Systems},
  33:12022--12033, 2020.

\bibitem{deng2022temporal}
Shikuang Deng, Yuhang Li, Shanghang Zhang, and Shi Gu.
\newblock Temporal efficient training of spiking neural network via gradient
  re-weighting.
\newblock {\em arXiv preprint arXiv:2202.11946}, 2022.

\bibitem{ding2023improved}
Yongqi Ding, Lin Zuo, Kunshan Yang, Zhongshu Chen, Jian Hu, and Tangfan Xiahou.
\newblock An improved probabilistic spiking neural network with enhanced
  discriminative ability.
\newblock {\em Knowledge-Based Systems}, 280:111024, 2023.

\bibitem{krizhevsky2009learning}
Alex Krizhevsky, Geoffrey Hinton, et~al.
\newblock Learning multiple layers of features from tiny images.
\newblock 2009.

\bibitem{Amir2017lowpower}
Arnon Amir, Brian Taba, David Berg, Timothy Melano, Jeffrey McKinstry, John
  DiBenedetto, Alexander Andreopoulos, Michael Chang, Efraim DeNolf, Taposh
  Nayak, et~al.
\newblock A low power, fully event-based gesture recognition system.
\newblock In {\em Proceedings of the IEEE Conference on Computer Vision and
  Pattern Recognition (CVPR)}, pages 7243--7252. IEEE, 2017.

\bibitem{hodgkin1952quantitative}
Alan~L Hodgkin and Andrew~F Huxley.
\newblock A quantitative description of membrane current and its application to
  conduction and excitation in nerve.
\newblock {\em The Journal of physiology}, 117(4):500, 1952.

\bibitem{lapicque1907recherches}
L~Lapicque.
\newblock Recherches quantitatives sur l’excitation electrique des nerfs.
\newblock {\em J. Physiol. Paris}, 9:620--635, 1907.

\bibitem{izhikevich2003simple}
Eugene~M Izhikevich.
\newblock Simple model of spiking neurons.
\newblock {\em IEEE Transactions on neural networks}, 14(6):1569--1572, 2003.

\bibitem{gerstner2002spiking}
Wulfram Gerstner and Werner~M Kistler.
\newblock {\em Spiking neuron models: Single neurons, populations, plasticity}.
\newblock Cambridge university press, 2002.

\bibitem{vaswani2017attention}
Ashish Vaswani, Noam Shazeer, Niki Parmar, Jakob Uszkoreit, Llion Jones,
  Aidan~N Gomez, {\L}ukasz Kaiser, and Illia Polosukhin.
\newblock Attention is all you need.
\newblock {\em Advances in neural information processing systems}, 30, 2017.

\bibitem{bahdanau2014neural}
Dzmitry Bahdanau, Kyunghyun Cho, and Yoshua Bengio.
\newblock Neural machine translation by jointly learning to align and
  translate.
\newblock {\em arXiv preprint arXiv:1409.0473}, 2014.

\end{thebibliography}

\appendix

\section{Appendix}
\subsection{Leaky Integrate-and-Fire model}
Classic spiking neuron models include the H-H (Hodgkin-Huxley) model~\cite{hodgkin1952quantitative}, the LIF (Leaky Integrate-and-Fire) model\cite{lapicque1907recherches}, the Izhikevich model~\cite{izhikevich2003simple}, and the SRM (Spike Response Model)\cite{gerstner2002spiking}.
The LIF model has become the most widely used model among them. This model uses the charging and discharging behavior of an RC circuit to simulate a series of dynamic characteristics of neurons receiving signals and transmitting signals, and its iterative form can be expressed as follows
\begin{eqnarray}
    u_t^{i, n} &=& (1-\frac{1}{\tau}) * u_{t-1}^{i, n} * (1-s_{t-1}^{i, n}) + x_t^{i, n}
\end{eqnarray}
where $u_t^{i, n}$ denotes the membrane potential of the $i$-th neuron in the $n$-th layer at time $t$, $\tau$ is time constant, $s_{t-1}^{i, n}$  is the output spike of the neuron at the previous moment, and $x_t^{i, n}$ is the input received from the previous layer. This formula can also be expanded in detail into three processes: charge, fire, and reset. The charge process is as follows
\begin{eqnarray}
    u_t^{i, n} &=& (1-\frac{1}{\tau}) * u_{t-1}^{i, n} + x_t^{i, n}
\end{eqnarray}
The equation indicates that the membrane potential accumulation at time $t$ is composed of two parts: one is the residual potential after the membrane potential leakage at at the previous time step, and the other is the external input $x$ at the current time step.
Then the fire process can be described as
\begin{eqnarray}
    s_{t}^{i, n}=
        \begin{cases}
            1&  u_t^{i, n} \geq V_{th}, \\
            0&  otherwise. 
        \end{cases}
    \label{lif:fire}
\end{eqnarray}
where neurons produce a spike only when the membrane potential surpasses the threshold $V_{th}$. The final reset process can described as
\begin{eqnarray}
    u_t^{i, n} &=&  u_{t}^{i, n} * (1-s_{t-1}^{i, n})
\end{eqnarray}
in which the membrane potentials of neurons that have fired spikes are reset to resting potential.



\subsection{Efficient Channel Attention}
We used the ECA~\cite{wang2020eca}  mechanism to fuse signals from different layers when introducing interlaminar connections. Compared to most attention mechanisms\cite{vaswani2017attention,bahdanau2014neural}, this lightweight attention mechanism introduces very few parameters with one-dimensional convolution. Specifically, the ECA attention process first uses global average pooling on features to extract global information from each channel, then learns the dependencies between channels with one-dimensional convolution, and finally obtains weights with the sigmoid function. These processes can be represented as:
\begin{eqnarray}
    g(X) = Sigmoid(Conv_{1*1}(AvgPool(X)))
    \label{eq:eca}
\end{eqnarray}

\subsection{Experimental platform}
The experiments were conducted on a computing platform with the following hardware specifications, as shown in Table \ref{tab:hardware}.
\begin{table}[h]
\centering
\caption{Hardware Specifications of the Experimental Platform}
\label{tab:hardware}
\begin{tabular}{@{}ll@{}}
\toprule
\textbf{Component}      & \textbf{Specification}                             \\ \midrule
CPU           & Intel(R) Core(TM) i7-9700K CPU @ 3.60GHz \\ 
GPU                     & NVIDIA GeForce RTX 2080 Ti * 2                         \\ 
Memory                  & 64 GB DDR4 RAM                                   \\ 
Storage                 & 4 TB NVMe SSD                                     \\ 
Operating System        & Ubuntu 20.04.6 LTS                                  \\ 
Software                & Python 3.7, PyTorch 1.8.1, CUDA 11.2, cuDNN 8.1   \\\bottomrule
\end{tabular}
\end{table}

\subsection{Experimental setup for direct training}
We apply the proposed multi-bit information transmission mechanism to LIF neurons, introduce interlaminar connections into the Basic Block of ResNet, and obtain the ResNet-20 network, which we then use for direct training and testing of the network's performance. The hyperparameters of the each experiment are shown in the Tab.\ref{tab:hyperparameters1}.

\begin{table}[h]
\centering
\caption{Hyperparameters for static images and ablation experiment}
\label{tab:hyperparameters1}
\begin{tabular}{@{}cccccccc@{}}
\toprule
\textbf{LR} & \textbf{Batch Size} & \textbf{Optimizer} & \textbf{Weight Decay}  & \textbf{Epochs} & \textbf{Time Const}  &\textbf{Threshold} &\textbf{T}   \\ \midrule
0.1           & 64          & Sgd          & $10^{-4}$          & 100   &4 &0.6 &4           \\ \bottomrule
\end{tabular}
\end{table}

\begin{table}[h]
\centering
\caption{Hyperparameters for neuromorphic data}
\label{tab:hyperparameters2}
\begin{tabular}{@{}cccccccc@{}}
\toprule
\textbf{LR} & \textbf{Batch Size} & \textbf{Optimizer} & \textbf{Weight Decay}  & \textbf{Epochs} & \textbf{Time Const}  &\textbf{Threshold} &\textbf{T}   \\ \midrule
0.1           & 32          & Sgd          & $10^{-3}$          & 100   &4 &1 &40           \\ \bottomrule
\end{tabular}
\end{table}

\subsection{Experiments on ANN-SNN conversion}
We also test our multi-bit mechanism in ANN-SNN conversion, applying it to IF neurons. Here, instead of expanding the number of integer bits, we expand by three decimal bits, resulting in MBIF neurons that emit one integer and three decimal bits. Then, drawing on the code from \cite{bu2023optimal}, we build ResNet-18 and VGG-16 networks with these neurons and conduct classifications on the CIFAR-10 and CIFAR-100 datasets.

The results and comparisons with other works are shown in Tab.\ref{ANN-SNN}. On the CIFAR-10 dataset, our method achieves nearly zero loss on the VGG-16 network with only 64 time steps, while the baseline still has some loss at 128 time steps. With our method on the ResNet-18 network, we achieve a loss of 0.1$\%$ at 128 time steps, whereas the baseline still has over 0.70$\%$ loss. On the CIFAR-100 dataset, for both VGG-16 and ResNet-18, our method significantly outperforms the baseline at 128 time steps with 64 time steps. Furthermore, our method still retains some classification ability at ultra-low time steps 4, whereas the baseline completely loses its ability to recognize images at these steps.
\begin{table}[h]
    \centering
    \caption{Results of using the multi-bit mechanism in ANN-SNN conversion}
    \begin{tabular}{cccccccc}
        \toprule
        Dataset   &Network    & Neuron & T=4      & T=16    & T=64 & T=128    & ANN\\
        
        \midrule
         \multirow{4}{*}{CIFAR-10} & \multirow{2}{*}{VGG-16} &  IF  & -   & 49.27$\%$  & 94.01$\%$ & 94.65$\%$ & \multirow{2}{*}{95.00$\%$}\\  
           &   &  MBIF(1 + 3)  & -   & 93.69$\%$  & 95.00$\%$ & 94.98$\%$ &   \\ \cline{2-8} 
           & \multirow{2}{*}{ResNet-18}  & IF & - & 76.31$\%$  & 94.60$\%$ & 95.25$\%$ & \multirow{2}{*}{95.99$\%$} \\
           &                            & MBIF(1 + 3) & 80.62$\%$ & 94.50$\%$  & 95.78$\%$ & 95.89$\%$ &  \\ 

        \midrule
        \multirow{4}{*}{CIFAR-100} & \multirow{2}{*}{VGG-16} &  IF  & -  & 9.12$\%$ & 59.06$\%$ & 67.66$\%$ & \multirow{2}{*}{74.68$\%$}\\  
           &   &  MBIF(1 + 3)  & 15.9$\%$  &63.14$\%$ & 73.74$\%$ & 74.32$\%$ &  \\ \cline{2-8} 
           & \multirow{2}{*}{ResNet-18}  & IF & - & 9.32$\%$  & 62.41$\%$ & 71.40$\%$ & \multirow{2}{*}{77.55$\%$} \\
           &                            & MBIF(1 + 3) & 37.86$\%$ & 69.88$\%$  & 76.42$\%$ & 77.06$\%$ &  \\

        \bottomrule
    \end{tabular}
    \label{ANN-SNN}
\end{table}

\end{document}